\documentclass[letterpaper]{article} 
\usepackage{aaai25}  
\usepackage{times}  
\usepackage{helvet}  
\usepackage{courier}  
\usepackage[hyphens]{url}  
\usepackage{graphicx} 
\usepackage{xcolor}
\usepackage{natbib}  
\usepackage{caption} 
\usepackage{multirow} 
\usepackage{bbding}
\usepackage{utfsym}
\usepackage{pifont}

\usepackage{array}  
\usepackage{colortbl}
\usepackage{algorithm}
\usepackage{algorithmic}
\usepackage{subfigure}
\usepackage{amsmath}
\usepackage{amssymb}
\usepackage{newfloat}
\usepackage{listings}
\DeclareCaptionStyle{ruled}{labelfont=normalfont,labelsep=colon,strut=off} 
\floatstyle{ruled}
\newfloat{listing}{tb}{lst}{}
\floatname{listing}{Listing}
\title{Towards Open-Vocabulary Remote Sensing Image Semantic Segmentation}
\author{
	Chengyang Ye,
	Yunzhi Zhuge,
	Pingping Zhang\thanks{Corresponding author (zhpp@dlut.edu.cn).},
}
\affiliations{
	School of Future Technology, School of Artificial Intelligence, Dalian University of Technology\\
	
	yecy@mail.dlut.edu.cn, \{zgyz,zhpp\}@dlut.edu.cn
}

\usepackage{bibentry}

\begin{document}

\maketitle

\begin{abstract}
Recently, deep learning based methods have revolutionized remote sensing image segmentation.
However, these methods usually rely on a predefined semantic class set, thus needing additional image annotation and model training when adapting to new classes.
More importantly, they are unable to segment arbitrary semantic classes.
In this work, we introduce Open-Vocabulary Remote Sensing Image Semantic Segmentation (OVRSISS), which aims to segment arbitrary semantic classes in remote sensing images.
To address the lack of OVRSISS datasets, we develop LandDiscover50K, a comprehensive dataset of 51,846 images covering 40 diverse semantic classes.
In addition, we propose a novel framework named GSNet that integrates domain priors from special remote sensing models and versatile capabilities of general vision-language models.
Technically, GSNet consists of a Dual-Stream Image Encoder (DSIE), a Query-Guided Feature Fusion (QGFF), and a Residual Information Preservation Decoder (RIPD).
DSIE first captures comprehensive features from both special models and general models in dual streams.
Then, with the guidance of variable vocabularies, QGFF integrates specialist and generalist features, enabling them to complement each other.
Finally, RIPD is proposed to aggregate multi-source features for more accurate mask predictions.
Experiments show that our method outperforms other methods by a large margin, and our proposed LandDiscover50K improves the performance of OVRSISS methods.
The proposed dataset and method will be made publicly available at https://github.com/yecy749/GSNet.

\end{abstract}
%
\section{Introduction}
Remote sensing image analysis aims at processing and interpreting remote sensing images to provide insights into natural environments and human activities.
It serves as a pivotal tool in improving human welfare.
As a central technique, Remote Sensing Image Semantic Segmentation (RSISS) facilitates various real-world applications, including enhancing agricultural yields~\cite{RSI_for_agriculture}, mitigating natural disasters~\cite{RSI_for_disaster}, and managing land cover changes~\cite{yan2022fully,yan2023transy}.

Recently, deep learning has revolutionized RSISS by enabling automatic segmentation methods.
With the advantages of different networks, researchers have leveraged Fully Convolutional Networks (FCN)~\cite{FCN}, UNet~\cite{UNet}, and Vision Transformers (ViT)~\cite{ViT} to improve the RSISS performance.
Furthermore, there has been a growing interest in exploring partially supervised RSISS methods, such as few-shot methods~\cite{RSI_fewshot}, weakly supervised methods~\cite{RSISS_SparseAnno}, and semi-supervised methods~\cite{RSI_SemiSupervised}.
However, existing RSSIS methods cannot segment arbitrary semantic classes, since they train and test on a predefined set of classes.
\begin{figure}[t]
\centering
\includegraphics[width=1\columnwidth]{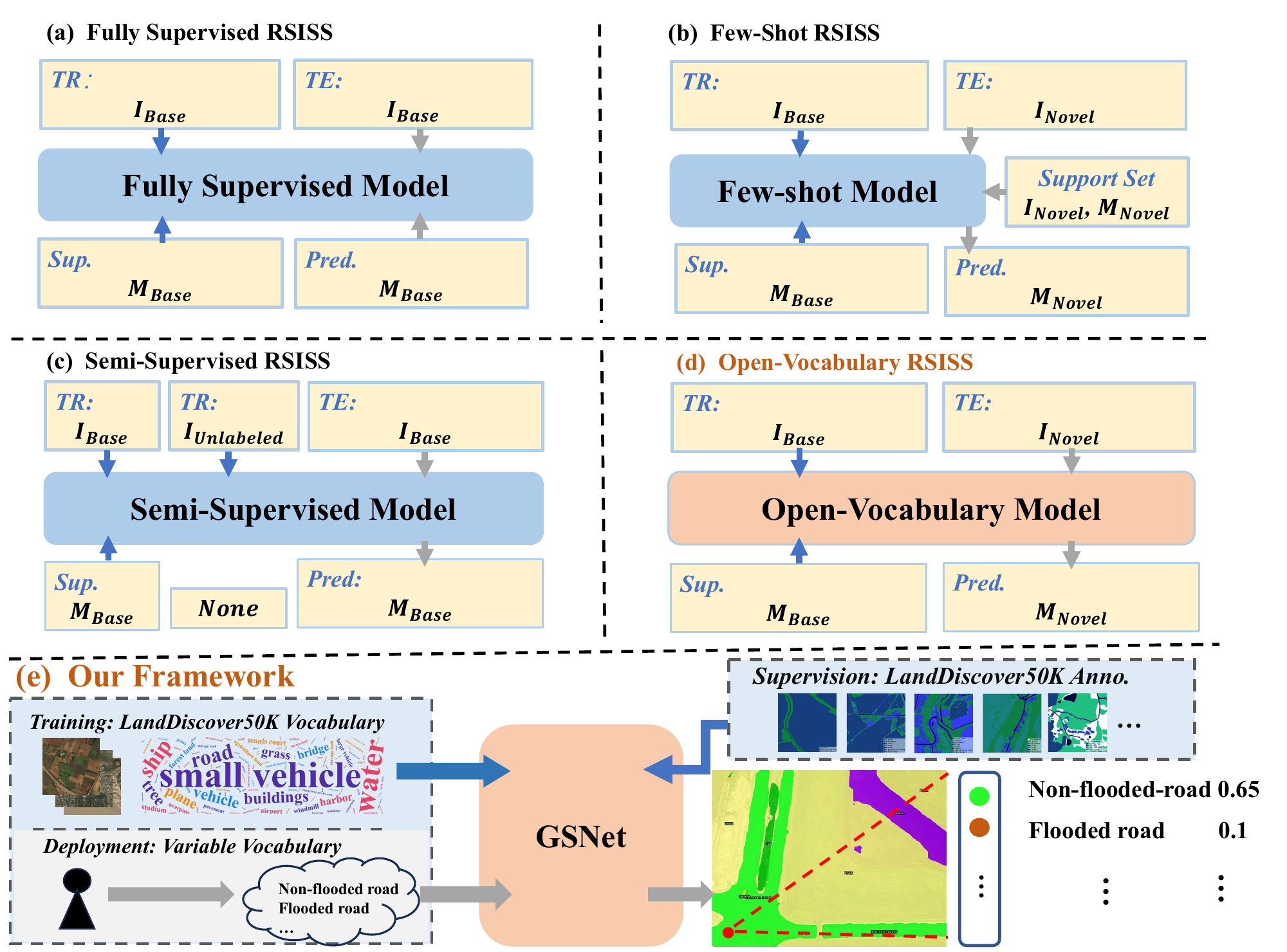}

\caption{Comparison of different learning paradigms for RSISS. (a) Fully supervised methods train and test on the same dataset. (b) Few-shot methods train on large annotated base classes and test on novel classes using a small support set. (c) Semi-supervised methods use large-scale unlabeled data with small-scale labeled base classes for training, then test on base classes. (d) Open-Vocabulary methods train on large-scale labeled data and test on arbitrary semantic classes. (e) Our framework is illustrated in brief.}

\label{fig:SettingComparison}
\end{figure}

With the advancement of RSI technologies, the excessive burden for data annotation highlights the need for a generalist model capable of adapting to diverse natural environments. Different from natural images, RSI often incurs significantly higher annotation costs due to its high resolution and inherent semantic ambiguity~\cite{RSI_ANNO}. However, state-of-the-art RSSIS methods are typically trained and tested on a predefined set of classes using annotated data, as shown in Fig. \ref{fig:SettingComparison} (a), (b), and (c). Considering the need for generalization, the above defects become even more significant. 
In fact, current RSISS approaches are not only unable to segment classes outside the predefined set but also struggle to generalize across different domains.

Furthermore, the importance of a generalist model in RSISS is amplified in scenarios requiring rapid responses, such as natural disasters, since there is insufficient time for extensive data annotation and model training. More specifically, due to the temporal and spatial changes in remote sensing data, great domain gaps exist in RSI. With such domain gaps, if a model can only perform accurate RSISS under specific conditions, its utility becomes severely restricted. Therefore, we propose Open-Vocabulary Remote Sensing Image Semantic Segmentation (OVRSISS) to address these challenges. OVRSISS aims at segmenting arbitrary semantic classes in RSI. Without the limitation of the pre-defined class set, OVRSISS allows users to flexibly switch between desired class sets based on their needs. Thus, OVRSISS not only reduces costs but also enables quicker responses in critical situations. We show the comparison between OVRSISS and existing RSISS learning paradigms in Fig. \ref{fig:SettingComparison}.

To the best of our knowledge, this is the first study to address the challenges of OVRSISS. Recognizing the absence of datasets specifically designed for OVRSISS, we developed LandDiscover50K, a dataset comprising 51,846 remote sensing images spanning 40 distinct classes. Alongside this dataset, we have formulated a comprehensive benchmark to facilitate the robust evaluation of OVRSISS methodologies.

For OVRSISS methods, there are two kinds of intuitive methods. First, one can simply train existing Open-Vocabulary Natural Image Semantic Segmentation (OVNISS) methods on LandDiscover50K. However, this kind of methods face notable performance limitations, primarily due to the absence of tailored designs for the RSI domain. Second, one can enhance existing OVNISS methods for RSI domain by replacing the generic CLIP~\cite{CLIP_origin} with domain-specific RemoteCLIP~\cite{remoteCLIP}. However, performance degradation is observed, primarily due to the limited generalization of RemoteCLIP. Both failures highlight the challenge of balancing domain-specific knowledge with generalization. Thus, how to effectively integrate RSI domain priors while maintaining a strong generalization ability remains an unsolved problem.

To address the above challenges, we propose GSNet, a novel framework tailored to effectively integrate RSI specialist domain priors with generalist CLIP. It employs a Dual-Stream Image Encoder (DSIE) to simultaneously extract generalist features from CLIP in parallel with RSI domain-specific features from a RSI backbone. A Query-Guided Feature Fusion (QGFF) is further introduced to integrate special RSI features and general features, enabling them to complement each other with the guidance of variable vocabularies. We also design a Residual Information Preservation Decoder (RIPD) to aggregate multi-source features for more accurate mask predictions.

To summarize, our contributions are as followed:
\begin{itemize}
\item We put forward Open-Vocabulary Remote Sensing Image Semantic Segmentation along with a tailored dataset named LandDiscover50K.
\item We propose a novel framework GSNet for OVRSISS which first extracts both generalist and specialist features with DSIE, followed by QGFF for multi-source feature fusion, and finally employs RIPD for information preservation and detail refinement. 
\item We conduct extensive experiments to demonstrate that our GSNet outperforms other state-of-the-art OVNISS methods by a large margin, and our LandDiscover50K markedly boosts the performance of OVRSISS methods.
\end{itemize}
\section{Related Work}
\subsection{Remote Sensing Image Semantic Segmentation}
Existing RSISS methods mostly focus on close-set performance evaluated on certain benchmarks. Datasets such as LoveDA~\cite{loveda}, iSAID~\cite{isaid} are proposed, enabling robust model training and evaluation. Based on these datasets, works such as FarSeg~\cite{FarSeg} and AerialFormer~\cite{AerialFormer} have successfully extended UNet and ViT to RSISS, boosting the performance on those datasets.

More recently, some work has addressed the data scarcity in RSI domain. For instance, Li et al.~\cite{RSI_SemiSupervised} delve into semi-supervised RSISS with consistency self-training. Jiang et al.~\cite{RSI_fewshot} use prototype-based semantic matching and a non-parametric metric learning loss to address few-shot RSISS. Hua et al.~\cite{RSISS_SparseAnno} propose a feature and spatial relational regularization to boost the performance of weakly-supervised RSISS. Zhu et al.~\cite{RSISS_UDA} delve into universal domain adaptation in RSISS, addressing domain distribution discrepancies through adversarial learning. However, none of them can segment arbitrary classes. 

As for RSISS datasets, we refer the readers to Tab.~\ref{tab:RSISS_Datasets} for more details. To our knowledge, there exists no prior dataset tailored for OVRSISS. Except for the recently proposed FLAIR~\cite{FLAIR} and SAMRS, most existing datasets are too small for OVRSISS. As for FLAIR, it only annotates regular land cover types, overlooking small objects, making it hard to smoothly adapt to OVRSISS settings. SAMRS~\cite{SAMRS} has successfully adapted several large-scale bounding-box annotated RSI object detection datasets to pixel-wisely annotated RSISS datasets. However, it has a limited generalization ability to land-cover segmentation tasks, which is widely used in multiple applications of RSI analysis.
\begin{table*}[t]
\centering
\setlength{\tabcolsep}{1.4mm}
\begin{tabular}{lllllllll}
	\hline
	\textbf{Datasets} & \textbf{Year} & \textbf{Images} & \textbf{Category} & \textbf{GSD (m)} & \textbf{Image Width} & \textbf{Total Area (km$^2$)} \\ \hline
	ISPRS Potsdam$^\ast$~\cite{ISPRS_Potsdam} & 2013 & 38 & 6 & 0.05 & 6,000 & 3.42 \\
	ISPRS Vaihingen~\cite{ISPRS_Vaihingen} & 2013 & 33 & 6 & 0.09 & 2,494 & 1.4 \\
	Zuric Summer~\cite{ZuricSummer} & 2015 & 20 & 8 & 0.62 & 1,150 & 9.37 \\
	UAVid~\cite{UAVid}& 2018 & 420 & 8 & - &3,840-4,096 & - \\
	Deep Globe Land Cover$^\dagger$~\cite{deepglobe} & 2018 & 1,146 & 7 & 0.5 & 2,448 & 1,716.9 \\
	iSAID~\cite{isaid} & 2019 & 2,806 & 15 & - & 800-13,000 & -\\
	FloodNet$^\ast$~\cite{floodnet} & 2020 & 3,200 & 9 & 0.02 & 3,000 & -\\
	GID~\cite{GID} & 2020 & 150 & 5 & 1-4 & 7,200 & 75,900 \\
	Landcover.ai~\cite{LandCover.ai} & 2020 & 41 & 3 & 0.25-0.5 &  800-13,000 & 216.27 \\
	LoveDA$^\dagger$~\cite{loveda} & 2021 & 5,987 & 7 & 0.3 & 1,024& 536.15 \\
	FLAIR$^\ast$~\cite{FLAIR} & 2022 & 77,762 & 19 & 0.1-0.2 &512 & 817 \\
	Open Earth Map$^\dagger$~\cite{oem} & 2022 & 5,000 & 8 & 0.25-0.5 & 1,024 & 799 \\
	SIOR$^\dagger$~\cite{SAMRS} & 2023& 23,463 & 20 & 0.5-30 & 800 & -\\
	SOTA$^\dagger$~\cite{SAMRS} & 2023 & 17,480 & 18 & - & 1,024 & -\\
	FAST$^\ast$~\cite{SAMRS} & 2023 & 64,147 & 37 & 0.3-0.8 & 600 & -\\
	\hline
	\textbf{LandDiscover50K} & 2024 & 51,846 & 40 & - & 400-1,200 & - \\
	\hline
\end{tabular}
\caption{Comparison of LandDiscover50K with other RSISS Datasets. Datasets marked with $\dagger$ contributed to the compilation of LandDiscover50K, while those marked with $\ast$ were selected for evaluating OVRSISS.}
\label{tab:RSISS_Datasets}
\end{table*}
\subsection{Open-Vocabulary Natural Image Semantic Segmentation}
OVNISS aims at segmenting arbitrary semantic class without needing explicit training examples for every possible category.
Recently, OVNISS has seen a significant performance boost thanks to the emergence of large-scale pre-trained Vision-Language Models (VLMs) such as CLIP.
There are mainly two lines of work in OVNISS, i.e., single-stage methods and two-stage methods.
For single-stage methods, LSeg~\cite{LSeg} leverages CLIP's patch embeddings, in line with CLIP's text embeddings to build the correlation of patches and texts.
SAN~\cite{SAN} introduces an additional backbone alongside a frozen CLIP, to directly generate and classify region proposals.
CAT-SEG~\cite{catseg} computes a patch-level cost volume map between image and text embeddings, and refines the cost volume map to obtain the final prediction.
Furthermore, SED~\cite{SED} enhances the performance by introducing a hierarchical encoder-decoder framework.
For two-stage methods, OpenSeg~\cite{OpenSeg} decouples the OVNISS task into class-agnostic proposal generation and proposal classification.
OVSeg~\cite{OVSeg} refines CLIP with mask-adapted image-text pairs to better classify mask proposals.
SCAN~\cite{SCAN} incorporates CLIP's generalized semantic prior into proposal embeddings to prevent collapsing on known categories and applies a contextual shift strategy to address global context gaps.
\section{LandDiscover50K Dataset}
\begin{figure}[t]
\centering
\includegraphics[width=0.95\columnwidth]{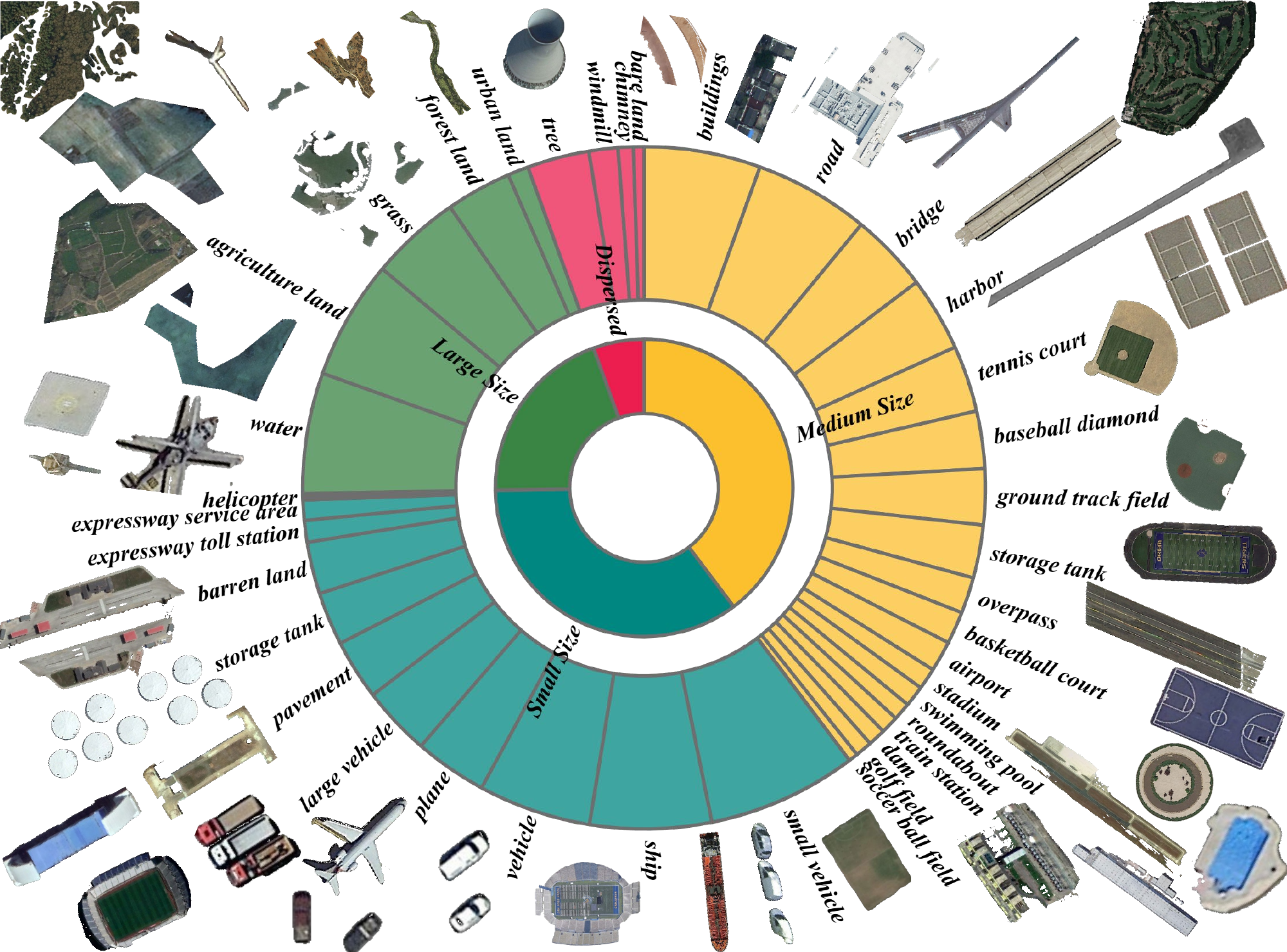} 
\caption{Illustration of the semantic class distribution and visual samples from LandDiscover50K. Sample images are strategically positioned adjacent to their corresponding semantic class tags for clarity. }
\label{fig:dsoverview}
\end{figure}
To address the lack of generalizable datasets on OVRSISS, we present LandDiscover50K. This dataset is designed to overcome several limitations in existing RSISS datasets. 
LandDiscover50K includes 51,846 meticulously selected high-resolution remote sensing images annotated across 40 object classes.
Meanwhile, LandDiscover50K addresses domain shift in RSI semantic segmentation by incorporating diverse sensors, resolutions, class variations, and ground sample distances. 
Moreover, LandDiscover50K can enhance the model robustness and generalization through the integration of fine-grained small-target datasets and large-scale land cover datasets.
\begin{figure*}[t]
\centering
\includegraphics[width=0.9\textwidth]{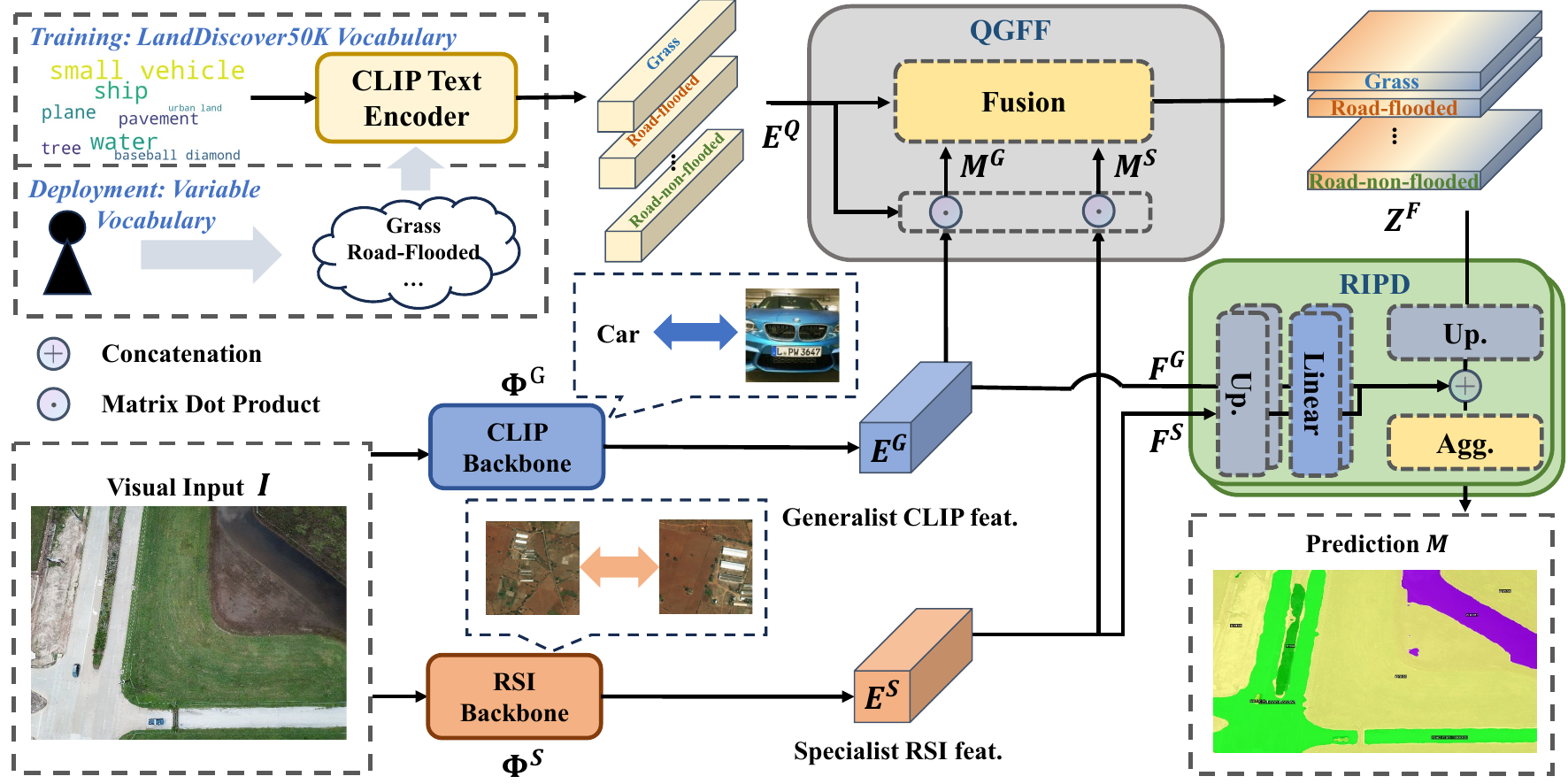} 
\caption{The overall architecture of GSNet. DSIE consists of a generalist CLIP backbone and a specialist RSI backbone. The specialist RSI backbone is pre-trained on RSI using self-supervised learning paradigm, while CLIP is pre-trained on image-text datasets using contrastive learning paradigm. QGFF enables dual stream features to complement each other under the guidance of variable vocabularies. RIPD further aggregates the multi-source features for more accurate mask predictions.}
\label{BigFrame}
\end{figure*}
\subsection{Data Acquisition and Annotation}
To build the LandDiscover50K dataset, we collect images from established RSISS datasets that provide pixel-wise annotations.
These datasets include Open Earth Map (OEM)~\cite{oem}, LoveDA~\cite{loveda}, Deep Globe Land Cover~\cite{deepglobe}, SIOR~\cite{SAMRS}, and SOTA~\cite{SAMRS}, with a detailed overview provided in Tab.~\ref{tab:RSISS_Datasets}.
LandDiscover50K comprises 51,846 image-annotation pairs across 40 diverse classes, drawn from sources such as DOTA~\cite{DOTA}, DIOR~\cite{DIOR}, xBD~\cite{xBD}, Inria~\cite{Inria}, OpenCities AI~\cite{OpenCitiesAI}, SpaceNet~\cite{SpaceNet}, LandCover.ai~\cite{LandCover.ai}, AIRS~\cite{AIRS}, GeoNRW~\cite{GeoNRW} and HTCD~\cite{HTCD} etc.
To ensure the comprehensive evaluation, fair comparison, and standardized preprocessing, we chose to utilize only the RGB modality, which is broadly accessible and aligns with the needs of open-vocabulary tasks.

For annotations, our approach emphasizes diversity and scale. We merge identical classes and preserve unique fine-grained ones.
In addition, we consolidate generic ``background'' labels into a single ``unlabeled'' category to mitigate over-fitting risks associated with the semantic bias introduced by the varying purposes of the source datasets.
\subsection{Statistics and Analysis}
As shown in Fig.~\ref{fig:dsoverview}, the LandDiscover50K dataset covers a diverse range of semantic classes from expansive land covers to salient objects. This diversity is crucial for modeling general variations inherent in real-world remote sensing tasks.
In addition, LandDiscover50K provides a balanced spatial coverage of segments within images, facilitating robust model training and reducing positional biases. Visualization of the attributes of LandDiscover50K are detailed in the Supplementary Materials.
\section{Our Method}
In this section, we introduce the Generalist and Specialist Network (GSNet) for OVRSISS.
The overall architecture of GSNet is depicted in Fig.~\ref{BigFrame}.
It employs a DSIE, integrating a domain-specific RSI specialist image encoder and a CLIP-based image-text aligned image encoder.
DSIE produce two kinds of complementary feature maps that are then harmonized using a Query-Guided Feature Fusion to leverage both generic and domain-specific strengths. 
The fused features are further denoised and upsampled through the RIPD, which ensures the preservation of critical information from both the specialist and generalist streams to generate the final segmentation results. 
We will delve into the detailed designs of GSNet in the following sections.
\subsection{Dual-Stream Image Encoder}
Directly applying CLIP to the RSI domain or using a RSI-specific model like RemoteCLIP has been found to under-perform on OVRSISS tasks. 
This is mainly because RemoteCLIP is pre-trained on a much smaller dataset compared to the general CLIP model.
In addition, the general CLIP lacks the specialized knowledge for OVRSISS. 
Thus, we propose a Dual-Stream Image Encoder that synergistically extracts specialist features and generalist features.
\subsubsection{Generalist CLIP Backbone.}
Following the most common practices, We use the ViT/B-16 architecture of the pre-trained CLIP. 
More specifically, we employ the image encoder of CLIP excluding the final projection heads, denoted as $\Phi^G$. 
Given an image $I \in \mathbb{R}^{H \times W \times 3}$, we extract the final feature $E^G \in \mathbb{R}^{H' \times W' \times D}$, where $D$ denotes the feature dimension and $\{H',W'\}=\{H/16, W/16\}$. With the query set $C_N$, we extract query embeddings $E^Q \in \mathbb{R} ^{N \times D}$ with CLIP text encoder $\Phi^{Q}$, as depicted in Fig.~\ref{BigFrame}.
\subsubsection{Specialist RSI Backbone.}
Although CLIP excels in recognizing novel objects, its effectiveness diminishes in dense segmentation tasks which are typical in complex remote sensing images.
This is primarily due to its lack of RSI domain priors.
Thus, we propose a specialist RSI backbone to enhance the RSI domain priors complementing the general CLIP model.
Specifically, it incorporates self-supervised pre-trained DINO~\cite{DINO}, which conducts contrastive learning utilizing local and global image views to capture spatial hierarchies effectively.
It effectively introduces RSI domain priors, while mitigating potential over-fitting that can be harmful to the generalization ability of CLIP.
Moreover, the specialist RSI backbone is trained solely on the image set of LandDiscover50K without labels, minimizing its latent impact on the generalization ability of GSNet.
Note that, DINO adopts the same ViT architecture as CLIP, ensuring a seamless integration.
With the specialist RSI backbone $\Phi^S$, we can extract the specific feature $E^S \in \mathbb{R}^{H' \times W' \times D}$.
\subsection{Query-Guided Feature Fusion}
One of the biggest challenges in OVRSISS is establishing robust associations between texts and images.
To address this issue, CAT-SEG~\cite{catseg} simply utilizes the cosine similarity of CLIP's text and image embeddings.
Differently, our approach does not rely solely on CLIP's image embeddings.
Instead, we decouple the scene features into specialist features and generalist features.
Then, we integrate text embeddings with both generalist CLIP's embedding and specialist RSI embedding.
Hence, our approach offers advantages over CAT-SEG in RSI domain, which will be further evaluated in the experiment section.
In addition, since the RSIB remains fixed during training, we do not introduce substantial computational overhead involved in training and testing.
Specifically, after obtaining $E^G $ and $E^S$, we normalize them and respectively compute matrix dot products with the embedded text-based query features, then we obtain the cost volume map of dual streams: $M^G$ and $M^S$, as follows:
\begin{equation}
M^{\{G, S\}}_{H',W',N} = \sum_D \text{norm}(E^{\{G, S\}}_{H',W',D}) \cdot \text{norm}((E^Q_{N,D})^\top),
\end{equation}
where \text{norm}($X$) denotes the $L_2$ normalization in channel dimension. This ensures that each query benefits from dual-stream image embeddings.
Afterwards, we further embed the volume map to a latent space:
\begin{equation}
Z^{\{G, S\}}_{H' \times W' \times N \times D} = \sigma(\varphi_{7 \times 7}(M^{\{G, S\}}_{H' \times W' \times N})),
\end{equation}
where $\varphi_{7 \times 7}$ denotes the convolution operation with a kernel size of $7 \times 7$, and $\sigma$ denotes the sigmoid activation function. 
The processed feature maps then undergo concatenation and aggregation, producing a query-guided fused feature map. To preserve the generalist integrity of CLIP and prevent feature degradation, a residual connection is applied to the fused feature map. The process is formed as follows:
\begin{equation}
Z^{F} = \sigma\left(\varphi_{7 \times 7}(Z^G \oplus Z^S)\right) + Z^G,
\end{equation}
where $\oplus$ denotes the concatenation operation in channel dimension, $+$ denotes the element-wise addition. 
Note that, in the above procedure, we use a simple structure design to validate our ideas.
%
More complex structures can be used to further improve the performances.
\subsection{Residual Information Preservation Decoder}
Since OVRSISS is a pixel-wise labeling task, prediction noises are inevitably introduced when using a versatile backbone.
Firstly, we utilize the ViT-B/16 as the image encoder, which downsamples images by a factor of 16, unavoidably resulting in some loss of details.
Secondly, the matrix dot product between image embeddings and text embeddings compresses information within the hidden space, further introducing semantic ambiguities.
To address the above challenges, we introduce RIPD for the backbone regularization and detail refinement, reducing prediction noises.
RIPD aggregates multi-source features for more accurate mask predictions.
Specifically, after obtaining the query-guided fused feature maps $Z^F$ from QGFF, we first adopt deconvolution to obtain $\tilde{Z}^F_0$. 
Then, the upsampled feature map $\tilde{Z}^F_0$ is concatenated with the projected intermediate feature maps $\tilde{F}_n^{\{G, S\}}$ from the CLIP backbone and RSI backbone.
Here, $n$ is the layer number. 
This is followed by a linear projection for dimension reduction. Then, we concatenate $Z^F$, $\tilde{F}_n^G$, and $\tilde{F}_n^S$ in channel dimension. Finally, we aggregate the concatenated feature to obtain $\tilde{Z}^F_1$. 
The whole RIPD is build by stacking the above block as follows:
\begin{equation}
 \tilde {F}^{\{G, S\}}_{n_i} = \text{Linear}(\mathcal{D}(F^{\{G, S\}}_{n_i})),
\end{equation}
\begin{equation}
\tilde {Z}_{i+1} = \text{AGR}(\text{AGR}(\varphi_{3 \times 3}(\mathcal{D}(\tilde{Z}_{i}) \oplus  \tilde {F}^G_{n_i}   \oplus \tilde {F}^S_{n_i}))),
\end{equation}
\begin{equation}
\text{AGR}(X) = \text{ReLU}(\text{GN}(\varphi_{3 \times 3}(X))),
\end{equation}
where $\mathcal{D}$ is the deconvolution with stride 2. 
GN is the group normalization
In practice, we stack two blocks for 4x upsampling, followed by a single convolution to obtain the final prediction $M$.
\section{Experiments}
\subsection{Datasets and Metrics}
\subsubsection{Datasets.}
Since there is no prior work, we select four representative datasets to assess the performance of OVRSISS: FLAIR~\cite{FLAIR}, FAST~\cite{SAMRS}, ISPRS Potsdam~\cite{ISPRS_Potsdam} and FloodNet~\cite{floodnet}. Detailed information about the testing datasets is presented in Tab.~\ref{tab:RSISS_Datasets} marked by *. Importantly, we do not restrict our evaluation to novel classes as a prerequisite for OVRSISS. We argue that cross-dataset validation is the most effective way to simulate real-world scenarios, as it reflects the complexity of real-world challenges while efficiently utilizing existing RSISS datasets. Among them, ISPRS Potsdam emphasizes in-vocabulary performance in OVRSISS with its high category similarity to the training set. FloodNet focuses on post-flood analysis, FLAIR covers large-scale land cover types, and FAST specializes in fine-grained object segmentation. These diverse datasets enable a comprehensive evaluation of the OVRSISS methods.
\subsubsection{Metrics.}
Following previous works~\cite{SimBaseline, OVSeg}, we employ the mean Intersection-over-Union (mIoU) as the segmentation metric for all experiments.
As for computation, we adopt GFLOPs and parameters. 
More details can be found in the Supplementary Materials.
\definecolor{morandiRed}{rgb}{0.93, 0.87, 0.87} 
\definecolor{morandiYellow}{rgb}{0.94, 0.92, 0.84} 
\definecolor{morandiBlue}{rgb}{0.89, 0.92, 0.95} 
\definecolor{morandiGray}{rgb}{0.94, 0.93, 0.91} 
\definecolor{morandiOrange}{rgb}{0.94, 0.84, 0.72} 
\begin{table*}[t]
\centering
\setlength{\tabcolsep}{3.1mm}
\begin{tabular}{lllllll}
	\hline
	Method & VLM & FLAIR & FAST & Potsdam & FloodNet & Avg. \\
	\hline
	\multicolumn{7}{l}{\textit{\textcolor[RGB]{100,100,100}{Fine-tuned on LandDiscover50K with the weight pre-trained on COCO}}}
	\vspace{1pt}
	\\
	CAT-SEG~\cite{catseg} & CLIP-ViT-B/16 & 18.94 & 15.23 & 36.01 & \underline{40.64} & \underline{27.71} \\
 	SCAN~\cite{SAN} & CLIP-ViT-B/16 & 19.19 & 10.02 & 5.68 & 36.61 & 17.87 \\
	OV-SEG~\cite{OVSeg} & CLIP-ViT-B/16 & 10.32 & 9.10 & 4.93 & 27.78 & 13.03 \\
	SAN~\cite{SAN} & CLIP-ViT-B/16 & 19.91 & 15.30 & 19.91 & 35.75 & 22.71 \\
	SED~\cite{SED} & CLIP-ConvNeXt-B & 18.59 & 12.90 & 28.16 & 31.65 & 22.83 \\
	EBSeg~\cite{EBSeg} & CLIP-ViT-B/16 & 21.26 & 18.53 & 5.68 & 35.26 & 20.18 \\
	\hline
	\multicolumn{7}{l}{\textit{\textcolor[RGB]{100,100,100}{Trained on LandDiscover50K}}}
	\vspace{1pt}
	\\
	CAT-SEG~\cite{catseg} & CLIP-ViT-B/16 & 19.99 & 13.90 & \underline{38.79} & 37.89 & 27.64 \\
    CAT-SEG~\cite{catseg}  & RemoteCLIP-ViT-B/32 & 14.36 &	10.79&	24.54&29.83&	19.88\\
	SCAN~\cite{SCAN} & CLIP-ViT-B/16 & 18.49 & 8.56 & 5.60 & 39.23 & 17.97 \\
    SCAN~\cite{SCAN} & RemoteCLIP-ViT-B/32 &15.38 &9.48 &10.15 & 19.05 & 13.51\\
    OV-SEG~\cite{OVSeg} & CLIP-ViT-B/16 & 13.65 & 7.47 & 5.02 & 15.17 & 10.33 \\
	SAN~\cite{SAN} & CLIP-ViT-B/16 & \textbf{22.48} & \underline{16.21} & 10.01 & 30.28 & 19.74 \\
	SED~\cite{SED} & CLIP-ConvNeXt-B & 14.65 & 12.63 & 28.64 & 22.57 & 19.62 \\
	EBSeg~\cite{EBSeg} & CLIP-ViT-B/16 & 20.24 & 15.15 & 8.37 & 37.93 & 20.42 \\
	\hline
	GSNet (Ours) & CLIP-ViT-B/16 & \underline{20.00} & \textbf{16.61} & \textbf{45.75} & \textbf{42.63} & \textbf{31.25} \\
	\hline
\end{tabular}
\caption{Performance comparison with other methods. The best and second-best results are in bold and underlined, respectively.}
\label{tab:big}
\end{table*}
\subsection{Implementation Details}
We train the GSNet with a per-pixel binary cross-entropy loss. Our implementation is based on PyTorch~\cite{pytorch} and Detectron2~\cite{detectron2}. We adopt AdamW~\cite{AdamW} as the optimizer, setting the learning rate to $2 \times 10^{-6}$ for the CLIP, while keeping the DINO fixed during the training process. The remaining parts of our model are randomly initialized and trained with a learning rate of $2 \times 10^{-4}$. The batch size is set to 4, and we use two NVIDIA RTX 3090 GPUs for training. We set the total number of training iterations to 30,000.
\subsection{Comparison with Other Methods}
Here, we compare our methods with recent state-of-the-art OVNISS methods~\cite{SAN,OVSeg,catseg,SED}. 
We re-evaluate all the methods to show their performance on OVRSISS. 
There are two settings: 1) fine-tuning on LandDiscover50K with the weight pre-trained on COCO-Stuff~\cite{COCO-Stuff} and 2) training exclusively on LandDiscover50K. 
For all methods, we deploy base models for a fair comparison.
\subsubsection{Quantitative Evaluation.}
Tab.~\ref{tab:big} presents the results of different methods on four datasets. 
It also shows the corresponding VLMs. 
For a fair comparison, we further improve the state-of-the-art methods SCAN and CAT-SEG by replacing CLIP with RemoteCLIP, which is pretrained on RSI domain. 
However, performance degradation occurs, further addressing the effectivenss of our proposed GSNet. 
Among these methods, our proposed method achieved the best averaged mIoU, outperforming the second best model by a large margin of 3.54\% mIoU.
\subsubsection{Qualitative Evaluation.}
\begin{figure}[t]
\centering
\includegraphics[width=0.9\columnwidth]{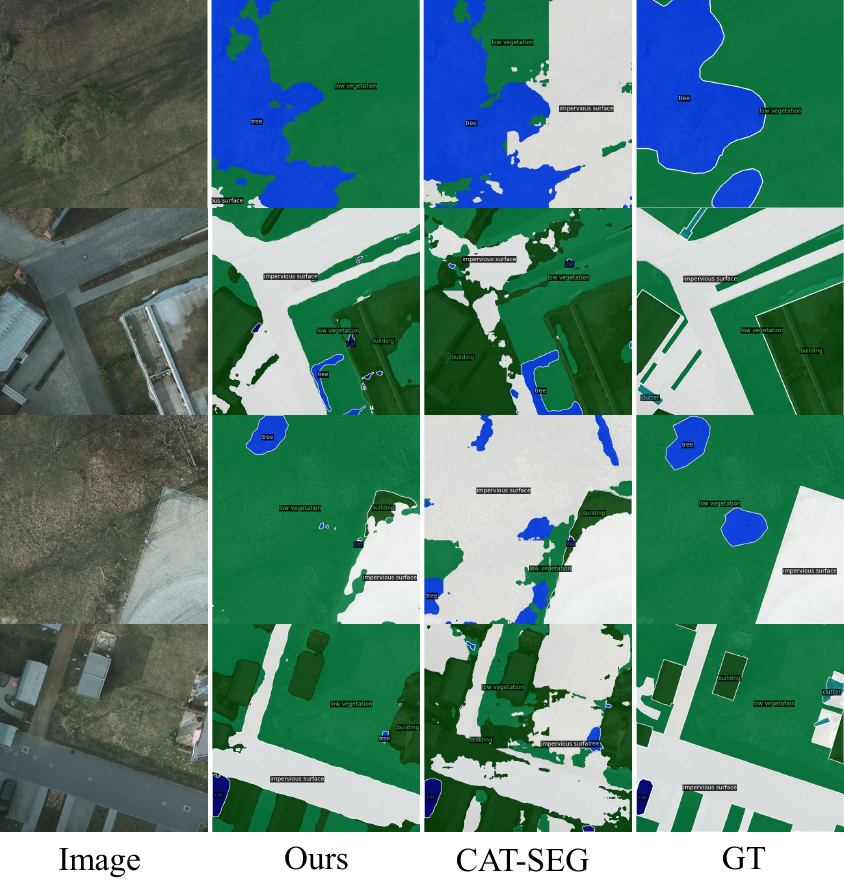}
\vspace{-2mm}
\caption{Qualitative evaluation of GSNet. Our method outperforms CAT-SEG in both semantic understanding and edge prediction.}
\vspace{-2mm}
\label{fig:VisComparison}
\end{figure}

Fig.~\ref{fig:VisComparison} shows the qualitative results of our proposed method versus CAT-SEG. Our method clearly exhibits superior boundary awareness and enhanced semantic recognition of target objects in these challenging remote sensing images.
\begin{figure}[t]
\centering
\includegraphics[width=0.96\columnwidth, height=0.74\columnwidth]{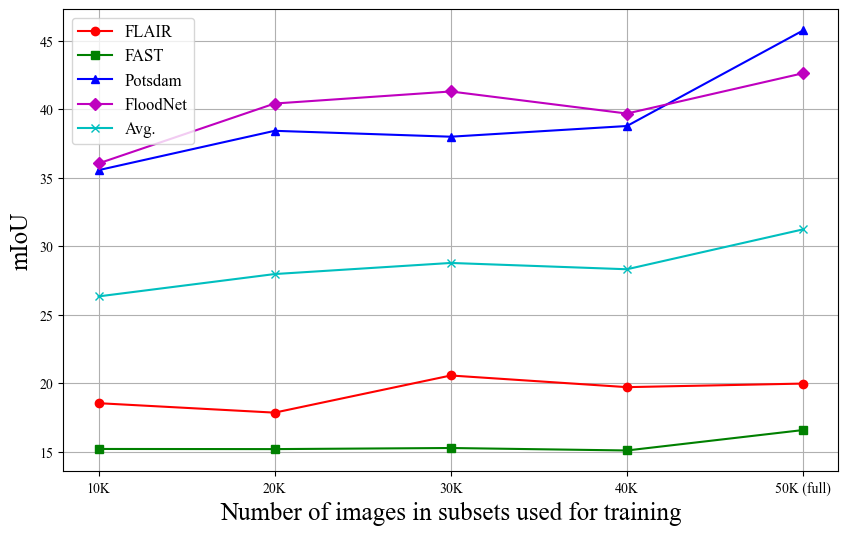}
\vspace{-2mm}
\caption{Performance of GSNet trained on different sizes of subsets of LandDiscover50K.} 
\vspace{-2mm}
\label{fig:SubSetsResults}
\end{figure}
\begin{table*}[t]
\centering
\setlength{\tabcolsep}{1.54mm}
\begin{tabular}{l>{\columncolor{morandiRed}}c>{\columncolor{morandiRed}}c>{\columncolor{morandiRed}}c|>{\columncolor{morandiYellow}}c>{\columncolor{morandiYellow}}c>{\columncolor{morandiYellow}}c|>{\columncolor{morandiBlue}}c>{\columncolor{morandiBlue}}c|lllll}

	\hline

	 & CLIP & RSI-CLIP & RSIB & Cat & Fusion Q & QGFF & VLMs & RSIB & \multicolumn{1}{c}{FLAIR} & \multicolumn{1}{c}{FAST} & \multicolumn{1}{c}{Potsdam} & \multicolumn{1}{c}{FloodNet} & \multicolumn{1}{c}{Avg.} \\
	
	\hline
	A&$\checkmark$& $\usym{2613}$ & $\usym{2613}$ & $\usym{2613}$ & $\usym{2613}$ & $\usym{2613}$ & $\checkmark$  &$\usym{2613}$ &   18.62 &   14.29 &   38.59 &   40.10 &   27.90 \\
	
	B&$\usym{2613}$ &$\checkmark$ &  $\usym{2613}$ & $\usym{2613}$ & $\usym{2613}$ & $\usym{2613}$  & $\checkmark$  &$\usym{2613}$ &   14.36 &   10.79 &   24.54 &   29.83 &   19.88 \\
	
	C&$\usym{2613}$ & $\usym{2613}$  &$\checkmark$ & $\usym{2613}$ & $\usym{2613}$ & $\usym{2613}$    &$\usym{2613}$ & $\checkmark$ &   3.87 &   2.63 &   14.65 &   6.74 &   6.98 \\
	
	D&$\usym{2613}$  &$\checkmark$ &$\checkmark$ &$\usym{2613}$  &$\usym{2613}$ &$\checkmark$ & $\checkmark$& $\checkmark$&   13.25&	  11.30&	  24.89&	  28.99&	  19.61 \\
	
	E&$\checkmark$  &$\usym{2613}$ &$\checkmark$ &$\checkmark$ & $\usym{2613}$  &$\usym{2613}$  & $\checkmark$& $\checkmark$&  14.47 &  10.18 &  18.57 &  32.37 &  18.90 \\
	
	F&$\checkmark$  &$\usym{2613}$ &$\checkmark$& $\usym{2613}$  & $\checkmark$  &  $\usym{2613}$   & $\checkmark$& $\checkmark$ &  8.84 &  10.12 &  24.32 &  35.92 &  19.81 \\
	
	G&$\checkmark$  &$\usym{2613}$ &$\checkmark$ & $\usym{2613}$  &$\usym{2613}$  & $\checkmark$  & $\usym{2613}$&$\usym{2613}$ &   17.70 &	  13.57&	  38.39&	  35.91 &   26.39 \\

	H&$\checkmark$  &$\usym{2613}$ &$\checkmark$ & $\usym{2613}$  &$\usym{2613}$  & $\checkmark$  & $\checkmark$&$\usym{2613}$ &   17.59 &     15.06 &   35.93 &   41.31 &   27.47 \\
	
	I&$\checkmark$  &$\usym{2613}$ &$\checkmark$ & $\usym{2613}$  &$\usym{2613}$  & $\checkmark$   &$\usym{2613}$ & $\checkmark$ &   17.06 &  15.30 &  36.50 &  33.09 &  25.49 \\
	\hline
	Ours&$ \checkmark$  &$\usym{2613}$ &$\checkmark$ &  $\usym{2613}$  & $\usym{2613}$  &  $\checkmark$   &  $\checkmark$ &    $\checkmark$ & 20.00 & 16.61 & 45.75 & 42.63 & 31.25 \\
	\hline
\end{tabular}
\caption{Ablation results on model components. Our GSNet is in the last row. Except for the ablated component, the rest of the architecture remains consistent with the full version. The red, yellow, blue parts respectively ablate the DSIE, QGFF, RIPD.}
\label{tab:components_ablation}
\end{table*}
\subsection{Ablation Studies}
\subsubsection{Ablation study on DSIE.}
The row A to D in Tab.~\ref{tab:components_ablation} present the ablation results on DSIE. 
We respectively adopt CLIP, Remote-CLIP, and RSIB (Specialist RSI Backbone) alone. 
In addition, we use RSI-CLIP to complement RSIB. 
Our proposed DSIE has outperformed previous methods by 12\%, 57\%, 348\%, and 59\% measured by average mIoU on four datasets.
Though pre-training on RSI image-caption pairs, RemoteCLIP's smaller annotated dataset leads to a weaker generalization on OVRSISS. 
Our RSIB alone under-performs, as it has not been exposed to any annotated data. 
Our proposed GSNet complements CLIP with RSIB's specialist priors, achieving a significant performance boost.
\subsubsection{Ablation study on QGFF.}
The row E to F in Tab.~\ref{tab:components_ablation} present ablation results on different designs of QGFF. 
The simple concatenation method, denoted as `Cat', first expands the text embeddings to the scale of image embeddings.
Then, it conducts feature concatenation on the two image embeddings with expanded text embeddings, followed by a linear projection to reduce the dimension.
This simple concatenation method lags behind our full version by a large margin of 65\%, showing the efficacy of QGFF. 
We further adopt an intuitive fusion and query design, denoted as `Fusion Q', which fuses the feature from two image encoders and then conducts the matrix dot product with query embeddings. 
This design also lags behind our proposed QGFF by 58\%.
\subsubsection{Ablation study on RIPD.}
The row G to I in Tab.~\ref{tab:components_ablation} present ablation results on various configurations of the RIPD. 
Compared with the non-integrated decoder, we achieved an improvement of 18\%. 
When integrating only CLIP or RSIB, the model further achieves improvements of 14\% and 23\%, respectively. 
The results further confirm the capability of RIPD to synergistically leverage both generalist features from CLIP and RSI domain priors from RSIB for more accurate mask predictions.
\subsubsection{Importance of LandDiscover50K.}
Our proposed LandDiscover50K paves the way of OVRSISS by introducing large-scale, multi-domain, multi-granularity RSI images with various class annotations. 
Here, we evaluate the importance of LandDiscover50K. 

First, we randomly sample 10,000, 20,000, 30,000, and 40,000 image-mask pairs from LandDiscover50K to create subsets. 
Then, we train GSNet on the subsets and other existing RSISS datasets, respectively. 
The results are presented in Fig.~\ref{fig:SubSetsResults}. As the scale of subsets increases, the performance is significantly improved. 
Compared with the smallest subset, the model achieves over 18\% performance improvement. 

We also train our model on typical RSISS datasets and validate them on the same datasets. 
As presented in Tab.~\ref{tab:AblationLandDiscover50K}, though the model trained on some RSISS dataset achieves impressive performance on specific test datasets, they consistently fall short compared to those trained on our proposed LandDiscover50K. 
This further validate the contribution of  our proposed dataset to OVRSISS.
\begin{table}[!ht]
\centering
\setlength{\tabcolsep}{1mm}
\begin{tabular}{llllll}
	\hline

	Training Set & FLAIR & FAST & Potsdam & FloodNet & Avg. \\
	\hline
	
	LoveDA & 8.92 & 10.32 & 18.87 & 7.67 & 11.45 \\
	DeepGlobe & 6.11 & 12.38 & \underline{21.26} & 11.56 & 12.82 \\
	SIOR & 13.26 & 13.93 & 13.80 & 14.47 & 13.87 \\
	SOTA &  12.91&  \textbf{18.70}&  14.94&  14.33&  \underline{15.22}
	\\
	OEM &  \underline{13.85}&  7.41&  17.76&  \underline{20.09}&  14.78
	\\
	\hline
	Ours &\textbf{20.00}& \underline{16.61} & \textbf{45.75} & \textbf{42.63} & \textbf{31.25} \\
	\hline
\end{tabular}
\caption{Ablation results with different training datasets.}
\label{tab:AblationLandDiscover50K}
\end{table}
\section{Conclusion}
In this work, we put forward the OVRSISS task and provide a large-scale dataset named LandDiscover50K for model training and testing. In addition, we develop GSNet, a novel framework that integrates RSI specialist domain priors with generalist pre-trained VLMs. Extensive experiments have evaluated the effectiveness of our methods and the dataset. We believe that this work will advance the current state of remote sensing research.
\section{Acknowledgments}
This work was supported in part by the National Natural Science Foundation of China (No.62101092).
\bibliography{aaai25}

\appendix
\newpage
\renewcommand{\thesection}{\Alph{section}} 
\renewcommand{\thesubsection}{\thesection.\arabic{subsection}} 

\setcounter{secnumdepth}{2} 
	\section{Introduction}
	The supplementary materials provide further validation of our proposed method and dataset. 
	Specifically, the supplementary material include:
	\begin{itemize}
		\item More method details
		\begin{itemize}
			\item Workflow of Specialist RSI Backbone
			\item More details for RIPD
			\item Computational complexity comparison
		\end{itemize}
		\item More dataset details
		\begin{itemize}
			\item Dataset attributes visualizations
			\item Dataset samples visualizations
			\item More details for evaluation datasets
		\end{itemize}
		\item More implementation details
		\begin{itemize}
			\item Text prompt templates
			\item Implementation details for RemoteCLIP
			\item Training and inference details
		\end{itemize}
		\item More experiment results
		\begin{itemize}
			\item Analysis of partial freezing strategy
			\item More qualitative results
		\end{itemize}
	\end{itemize}
	
	\section{More Details for GSNet}
	In this section, we provide additional details on the design and implementation of GSNet.
	
	\subsection{Specialist RSI Backbone}\label{RSIB} 
	In this paragraph, we further illustrate the RSIB, which is designed to extract specialist RSI domain priors. 
	For the training of RSIB, we follow the DINO~\cite{DINO} paradigm. 
	Specifically, we train RSIB on the image set of LandDiscover50K for 300 epochs, with a batch size of 8. 
	We adopt a teacher temperature of 0.07, with warmup epochs of 30. 
	For inference, only the teacher branch is utilized. 
	RSIB employs the architecture of ViT-B/8, resulting in a resolution of $\{H',W'\}=\{H/8, W/8\}$ and a channel dimension of $768$.
	To align with the CLIP backbone, we simply use a simple convolution of kernel size 2, stride 2.
	
	\subsection{Residual Information Preservation Decoder} 
	For RIPD, we adopt a residual strategy for information preservation. 
	Given RSIB has a different resolution of image features, we have: $F^S_{n_i} \in \mathbb{R}^{H/8, W/8, D'}$ and $F^G_{n_i} \in \mathbb{R}^{H/16, W/16, D}$, where $D'=768$ and $D = 512$. 
	For feature alignment and model symmetry, we adopt:
	\begin{equation}
		\tilde{F}^{G}_{n_1} = \text{Linear}(F^G_{n_1}) ,\quad
		\tilde{F}^{S}_{n_1} = \mathcal{D}_2(F^S_{n_1})
	\end{equation}
	and 
	\begin{equation}
		\tilde{F}^{G}_{n_2} = \mathcal{D}_2(F^G_{n_2}) ,\quad
		\tilde{F}^{S}_{n_2} = \mathcal{D}_4(F^S_{n_2}),
	\end{equation}where $\mathcal{D}_{i}$ denotes deconvolution of stride $i$, kernel size $i$. 
	Specifically, we use $\{n_1, n_2\} = \{4, 8\}$. 
	Thus, the dual stream backbone feature introduced to RIPD will be properly aligned.

	\begin{figure}[t]
		\centering
		\includegraphics[width=0.45\textwidth]{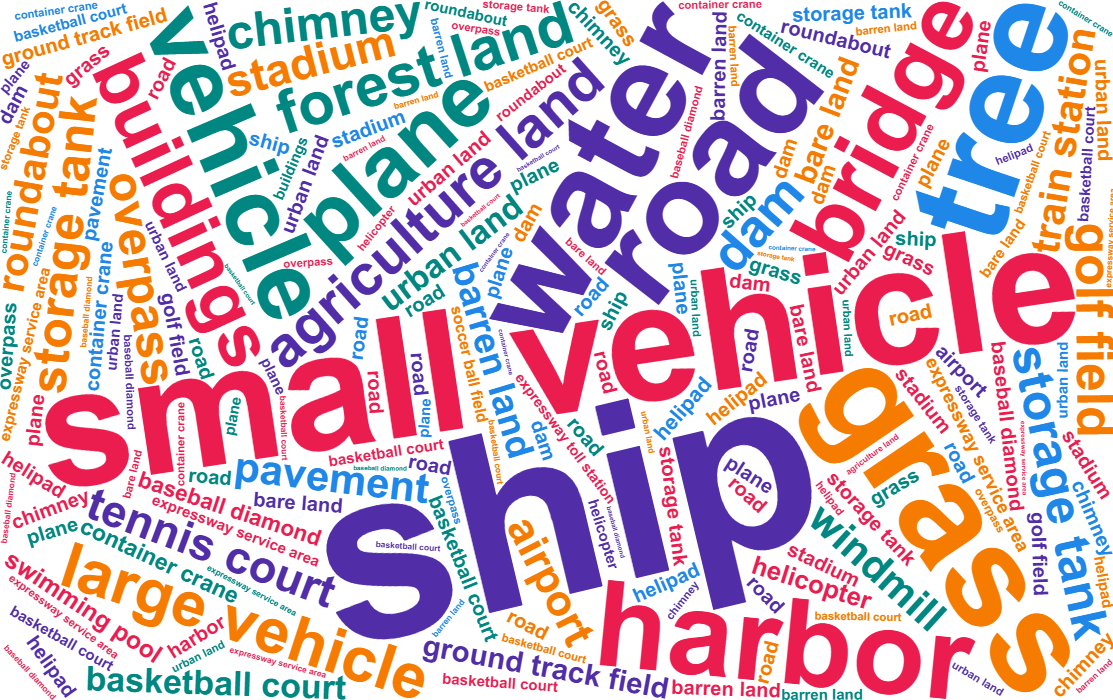}
		\caption{Word cloud of vocabulary in LandDiscover50K. We illustrate the most frequently occurring terms within the dataset. The size of each word corresponds to its frequency.}
		\label{fig:wordcloud}
	\end{figure}
	
	\subsection{Differences between GSNet and CAT-SEG}

	GSNet has three key modules, i.e., DSIE, QGFF and RIPD. 
	DSIE combines a RSI domain-specific backbone with CLIP to produce complementary yet unaligned dual-stream image features. 
	In contrast, CAT-SEG employs a single-stream CLIP visual encoder for feature extraction. 
	QGFF aligns and fuses features obtained from DSIE, while CAT-SEG only fuses image and text features without any alignment. 
	RIPD aggregates fused features with dual-stream residual connections, while CAT-SEG utilizes a simpler upsampling decoder. 
	The aforementioned technical differences establish GSNet as a new solution.

	Overall, our aim is to handle the OVRSISS task. 
	Thus, we effectively integrate domain-specific expertise of RSI with the generalist capacity of VLMs. 
	This new solution not only enhances OVRSISS performance but also verifies VLM' s capabilities in RSI tasks. 
	The CAT-SEG is only tested on natural images, lacking of generalization.
	\subsection{Computational Complexity}
	
	We compare the proposed method with the competitors as shown in Tab.~\ref{tab:model_comparison}. For fair comparison, we deploy base model and maintain the original inference settings for all methods. 
	\begin{table}[ht]
		\centering
		\begin{tabular}{llll}
			\hline
			\textbf{Model} & \textbf{Params} & \textbf{Trainable Params} & \textbf{FLOPs} \\ 
			\hline
			CAT-SEG       & 154M  & 25M  & 660G  \\ 
			OVSeg         & 531M  & 410M & 8,500G  \\ 
			SAN           & 437M  & 9M   & 64G  \\  
			SCAN          & 890M  & 158M & 8,290G \\ 
			SED           & 180M   &90M & 192G \\
			EBSeg      &  262M & 27M & 312G \\
			\hline
			ours  & 243M  & 29M  & 2,160G  \\ 
			\hline
		\end{tabular}
		\caption{Comparison of models in terms of parameters, trainable parameters, and FLOPs.}
		\label{tab:model_comparison}
	\end{table}
	
	\section{More Details for LandDiscover50K}
	In this section, we provide additional details for LandDiscover50K.
	\begin{table*}[t]
		\centering
		\setlength{\tabcolsep}{6.5mm}
		\begin{tabular}{llccccc}
			\hline
			CLIP & RSIB & FLAIR & FAST & Potsdam & FloodNet & Average \\
			\hline
			Freeze & Full & 11.61 & 12.72 & 22.25 & 17.65 & 16.06 \\
			Freeze & Attention & 11.66 & 13.68 & 28.18 & 28.88 & 20.60 \\
			
			Freeze & Freeze & 16.02 & 12.66 & 29.32 & 30.42 & 22.10\\
			
			Full   & Full	& 9.62&	4.31&	21.10&	26.88&	15.48  			\\
			Full   &Attention &    15.77&	14.40&	38.87&	35.96&	26.25	\\
			Full   & Freeze & 18.07 & 10.54 & 25.37  & 32.95 & 21.73 \\

			Attention	&Full	&19.47	&14.40	&37.26	&37.04	&27.04\\
			Attention	&Attention	&19.93&	15.09&	39.25	&37.90	&28.04\\
			\hline
			Attention	&Freeze	&20.00&	16.61 &45.75	&42.63	&31.25 \\
			
			\hline
		\end{tabular}
		\caption{Ablation Results on Training Strategy for GSNet. `attention' means fixing all the parameters except for the attention projection layers for `query' and `value'. }
		\label{tab:abl}
	\end{table*}
	\begin{figure*}
		\centering
		\includegraphics[width=0.96\textwidth]{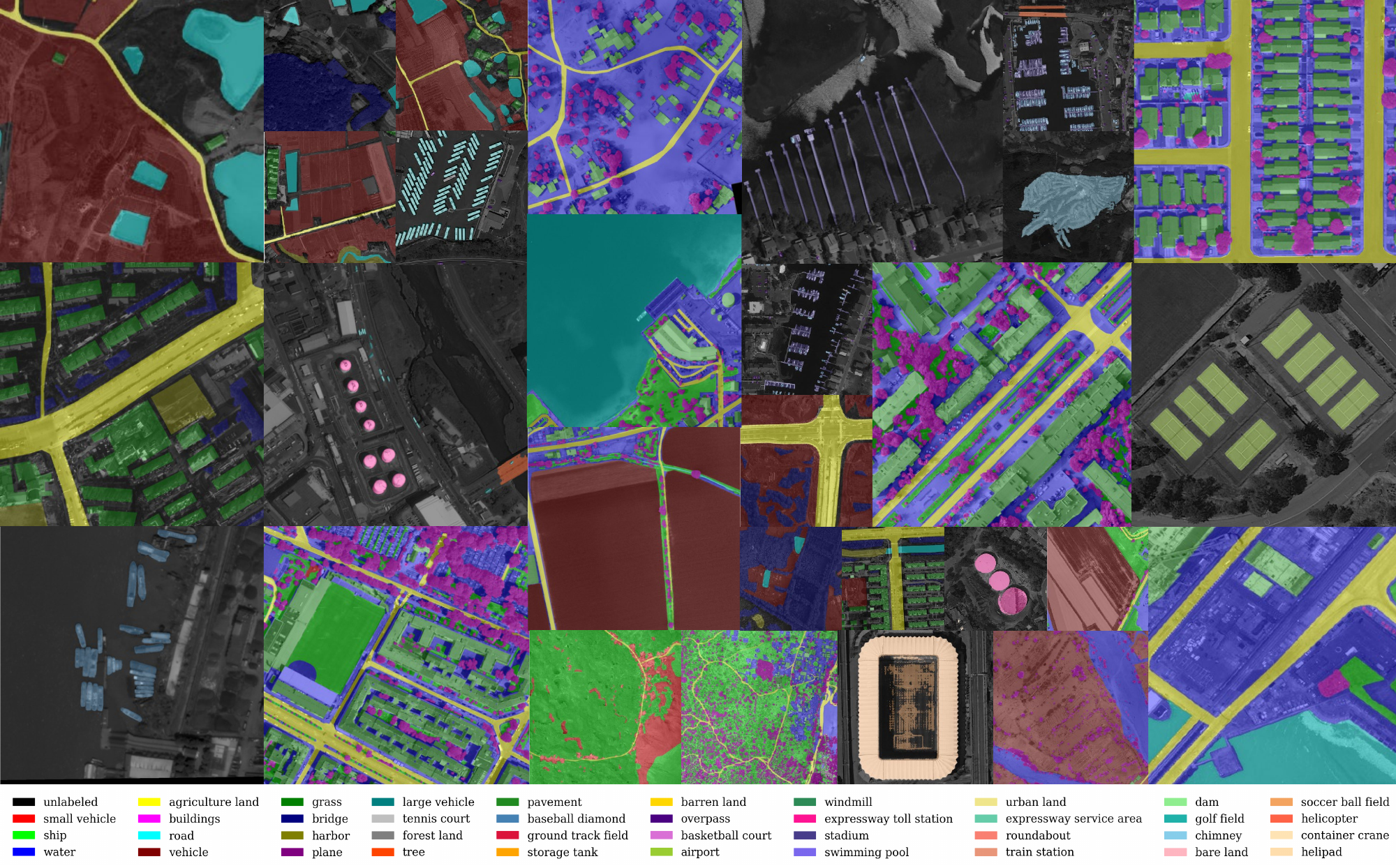}
		\caption{Samples of LandDiscover50K. Samples of LandDiscover50K. For better clarity, we convert the images to grayscale and overlay the labeled masks. The colors represent categories as indicated in the legend. }
		\label{fig:showLD}
	\end{figure*}
	\begin{figure*}
		\centering
		\includegraphics[width=0.96\textwidth]{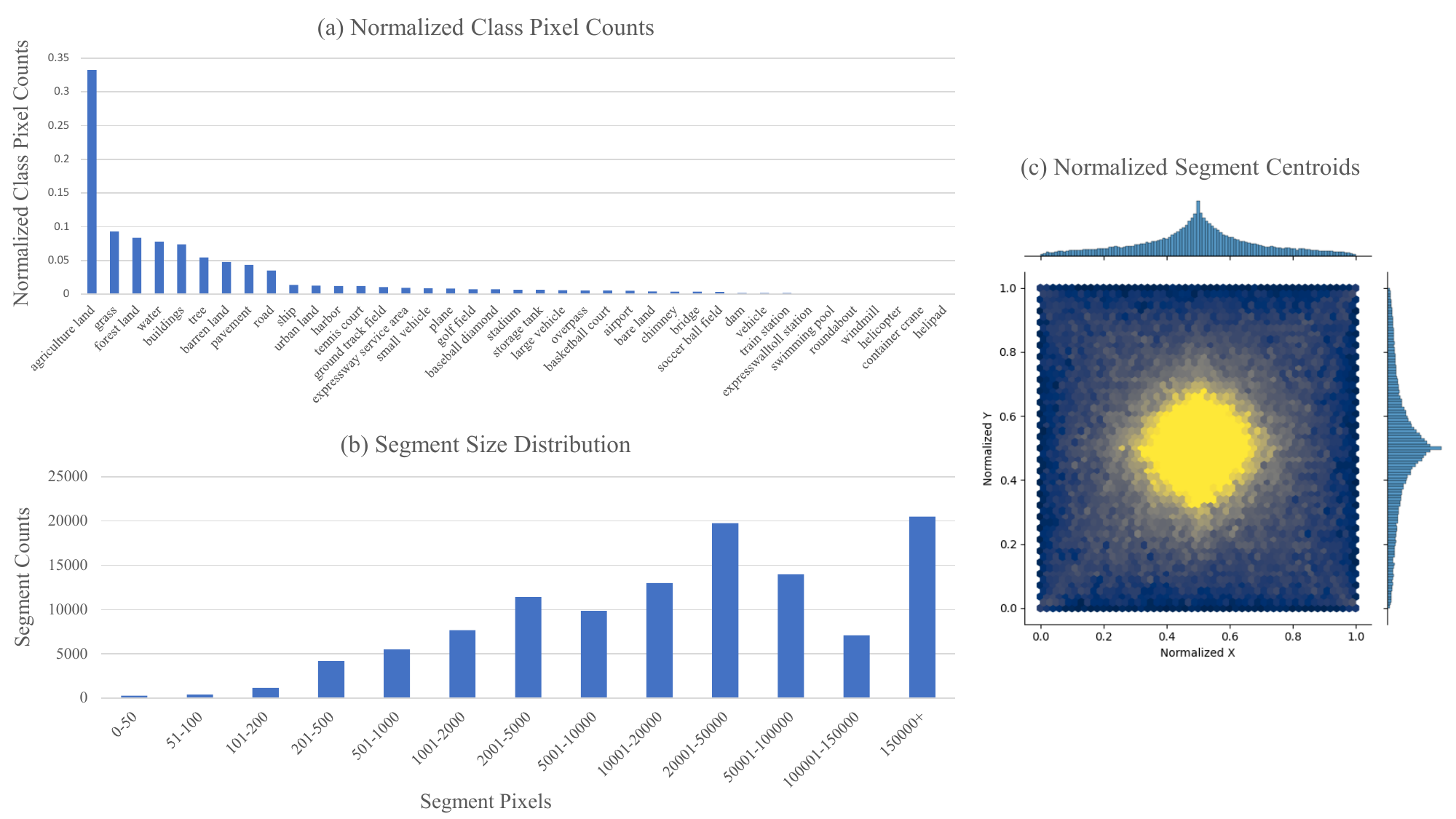}
		\caption{Visualization of attributes of LandDiscover50K. These attributes are detailed in section~\ref{statics}.}
		\label{fig:statistics}
	\end{figure*}

	\subsection{More Visualization of LandDiscover50K.}\label{statics} 
	For better illustration of LandDiscover50K, we provide word cloud visualization, attributes visualization and samples visualization in Fig.~\ref{fig:wordcloud}, Fig.~\ref{fig:showLD} and Fig.~\ref{fig:statistics} respectively.
	
	Specifically, in Fig.~\ref{fig:wordcloud}, we show the frequency of vocabulary appearing in the images within the datasets according to the size of each class. 
	For Fig.~\ref{fig:showLD}, a set of representative sample of LandDiscover50K are shown in grid. We indicate classes by the colored masks overlaid on the image, with the legend serving as a reference.	
	In Fig.~\ref{fig:statistics}, we use 3 statistical attributes. Normalized class pixel counts, segment size distribution, and normalized segment centroids. 
	Normalized class pixel counts represent the proportion of pixels for each class relative to the total number of pixels annotated. 
	Segment size distribution represents the frequency of different segment sizes (in pixels) within the dataset. 
	Normalized segment centroids represent the average central positions of segments within the dataset, normalized to a common scale.
	
	As shown in Fig.~\ref{fig:statistics} (a) and Fig.~\ref{fig:statistics} (b), LandDiscover50K encompasses a broad range of classes, including large objects like land cover types and small objects like vehicles and bridges. 
	This diversity in class annotations facilitates OVRSSS by establishing robust connections between images and texts based on large-scale data. 
	This is critical to the setting of OVRSSS, because in the real-world settings of RSI tasks, there exists great domain variation. 
	As visualized in Fig.~\ref{fig:statistics} (c), LandDiscover50K also offers a well-distributed and diverse spatial coverage of segment locations within images, ensuring robust model training by presenting segments in various positions and minimizing positional bias.

	\subsection{Evaluation Datasets}
	
	In this section, we explain the data pre-processing methods adopted for the evaluation datasets. 

	For ISPRS Potsdam, we split all high-resolution RSI images into $512 \times 512$ patches, following standard practice. We utilize all the data without further splitting, covering six common land cover types and 5472 patches. 
	For FLAIR, we only use the testing subset, consisting of 15,700 RSI images that represent 19 land cover types. 
	For FAST, we adopt the validation subset, which includes 3,207 RSI images that cover 37 categories of fine-grained ground objects. 
	For FloodNet, we use the validation set, which is aligned with the testing set and consists of 898 high-resolution UAV images across 9 real-world post-disaster land cover types.

	For all evaluation, we exclude the calculation of IoU for the generic `background' class throughout the evaluation process. This is due to the ambiguous definition of `background' class across datasets.
	
	Moreover, we calculate the label-set similarity between LandDiscover50K and the validation datasets, as shown in Tab.~\ref{tab:label_similarity}. We employ a plain CLIP ViT-B/16 text encoder to encode the label sets. Then, we calculate the Hausdorff distance for each comparison pair to measure the label-set similarity. 
	
	\section{More Implementation Details}
	
	\subsection{Text Prompt Templates.} 
	To ensure a fair comparison, we use the plain template `A photo of a \{class\}', without handcrafting the text prompts. 
	
	\subsection{Implementation Details for RemoteCLIP.} Since RemoteCLIP only offers pretrained model of ViT/B-32, we employ an upsampling strategy by increasing the image size to $768*768$ to maintain the feature size and secure parallel implementation. 
	
	\subsection{Training Details.} As discussed in the main body, we address the ambiguity surrounding the `background' class. 
	Given that LandDiscover50K integrates datasets with diverse purposes, the definition of `background' varies across them. 
	Therefore, we exclude the gradient of the `background' class during training across all methods. 
	This approach is consistently applied to all experiments conducted. 
	Moreover, for CLIP text encoder and image encoder, we adopt partial freezing, fixing all the parameters except for the attention projection layers for `query' and `value', as in CAT-SEG~\cite{catseg}. 
	And for RSIB, we freeze all the parameters.
	We further delve into the partial freezing strategy in Section~\ref{AddExp}.
	\subsection{Inference Details.} 
	\subsubsection{Patch Inference Strategy}
	To handle the high-resolution images in RSI, we adopt a patch inference strategy, similar in CAT-SEG~\cite{catseg}. 
	Specifically, we first reshape the input image to $640 \times 640$, then we split the reshaped image into 4 patches $384\times384$, with every adjacent two of them having an overlap of $384\times128$. 
	After obtaining all the patch inference results, we merge them together and get an averaged prediction. 
	\subsubsection{Inference on Novel Categories}
	As for inference on novel categories, we first input their category names into CLIP's text encoder to obtain the text embedding $E^Q$. 
	$E^Q$ can be treated as the query vectors. Then, we input $E^Q$ into the QGFF module. 
	With the dual-stream image features from CLIP backbone and RSI backbone, we obtain a query-based fused image feature specific to each query vector. 
	This fused image feature then serves as input to the RIPD decoder and upsampling process. 
	Finally, the framework produces the segmentation result based on each query. 
	Since the segmentation mask is based on text queries, the framework can segment arbitrary semantic class as long as it is specified.

	\begin{table}[t]
		\centering
		\setlength{\tabcolsep}{5mm} 
		\renewcommand{\arraystretch}{1.2} 
		\begin{tabular}{p{3cm}l}
			\hline
			\textbf{Dataset} & \textbf{Label-set Similarity} \\
			\hline
			FAST & 0.79 \\ %
			FLAIR & 0.69 \\ %
			Potsdam & 0.69 \\ %
			FloodNet & 0.71\\ %
			\hline
		\end{tabular}
		\caption{The label-set similarity between LandDiscover50K and validation datasets.}
		\label{tab:label_similarity}
	\end{table}	
	\section{Additional Experiment Results}\label{AddExp}
	\subsection{Analysis of Partial Freezing Strategy for GSNet}
	In this section, we additionally analyze the effects of partial freezing strategy. 
	We report the results of different fine-tuning strategy of RSIB and CLIP in Tab.~\ref{tab:abl}.
	Except for the DSIE, which is pre-trained on the image set of LandDiscover50K, the remainder of GSNet is initialized randomly. 
	Overall, attention fine-tuning for CLIP paired with frozen RSIB yields the best performance. 
	Freezing RSIB shows a large margin of 4.21 and 3.21 mIoU compared to full and attention fine-tuning. 
	We analyze that excessive fine-tuning on the semantically dense task of OVRSISS undermines RSIB's domain-specific expertise, given RSIB is densely trained under an unsupervised paradigm to acquire RSI domain knowledge.

	\subsection{More Qualitative Results}
\begin{figure*}
		\centering
		\includegraphics[width=0.8\textwidth]{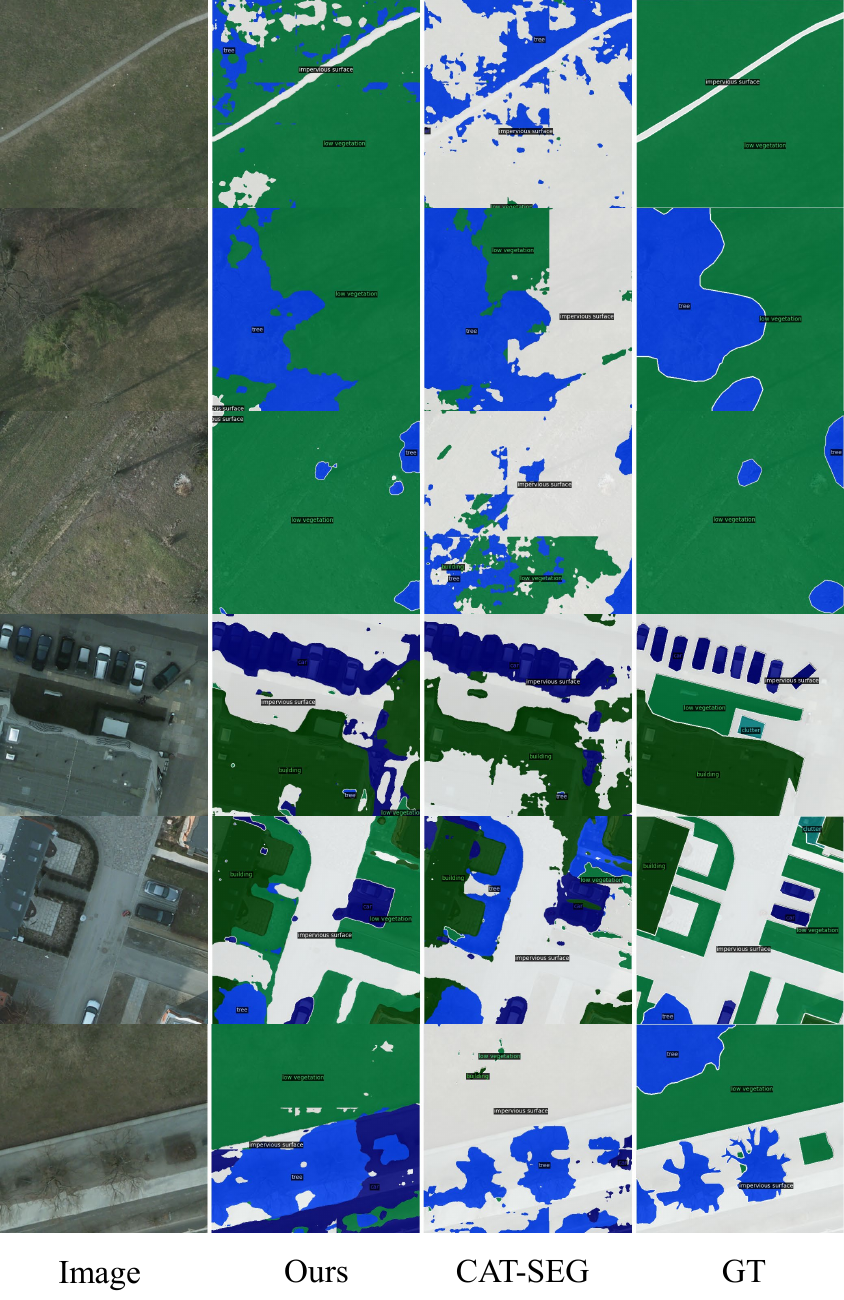}
		\caption{Qualitative results on ISPRS Potsdam~\cite{ISPRS_Potsdam}.}
		\label{fig:potsdam}
	\end{figure*}

\begin{figure*}
	\centering
	\includegraphics[width=0.8\textwidth]{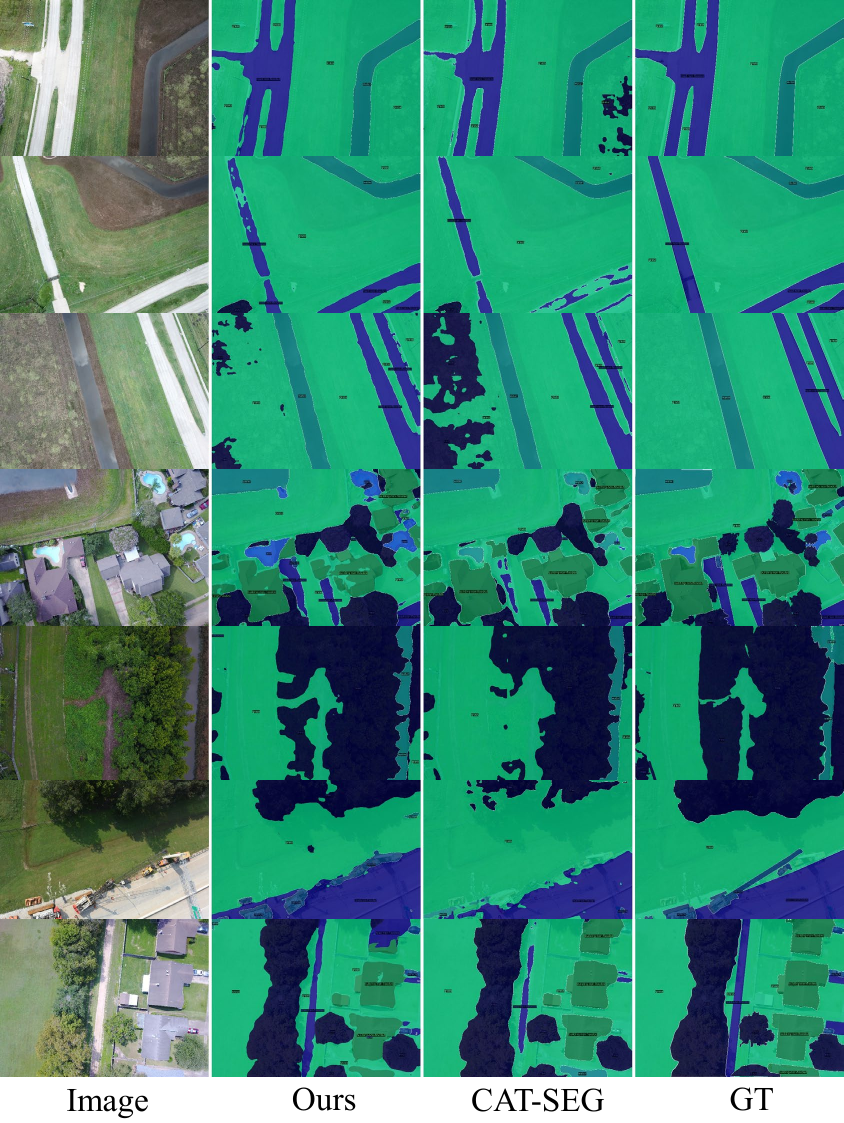}
	\caption{Qualitative results on FloodNet~\cite{floodnet}.}
	\label{fig:floodnet}
	\end{figure*}

	We provide more qualitative results on ISPRS Potsdam in Fig.~\ref{fig:potsdam}, FloodNet in Fig.~\ref{fig:floodnet}.

\end{document}